\pdfoutput=1

\documentclass[11pt]{article}

\usepackage[]{acl}

\usepackage{times}
\usepackage{latexsym}

\usepackage[T1]{fontenc}

\usepackage[utf8]{inputenc}

\usepackage{microtype}

\usepackage{inconsolata}
\usepackage{multirow}
\usepackage[spacesep,definevectors]{easyvector}
\usepackage{amsmath}
\usepackage{mathtools}
\usepackage{xspace}
\usepackage{algpseudocode}
\usepackage{algorithm}

\DeclareMathOperator*{\argmax}{argmax}
 
\newcommand{\inprod}[2]{\langle #1, #2\rangle} 

\newcommand{\OurMODEL}{ICE-NET} 
\newcommand{\eat}[1]{}

\title{Antonym vs Synonym Distinction using InterlaCed Encoder NETworks (\OurMODEL)}

\author{
    Muhammad Asif Ali,\textsuperscript{\rm 1} 
    Yan Hu,\textsuperscript{\rm 1}
    Jianbin Qin,\textsuperscript{\rm 2}
    Di Wang\textsuperscript{\rm 1}\\
    \textsuperscript{\rm 1} King Abdullah University of Science and Technology, KSA\\
    \textsuperscript{\rm 2} Shenzhen University, China\\
    \{muhammadasif.ali, yan.hu, di.wang\}@kaust.edu.sa, qinjianbin@szu.edu.cn\\
}

\begin{document}
\maketitle

\begin{abstract}
Antonyms vs synonyms distinction is a core challenge in lexico-semantic
analysis and automated lexical resource construction. 
These pairs share a similar distributional context which makes it harder 
to distinguish them.
Leading research in this regard attempts to capture the properties of the
relation pairs, i.e., symmetry, transitivity, and trans-transitivity.
However, the inability of existing research to appropriately model the 
relation-specific properties limits their end performance.
In this paper, we propose InterlaCed Encoder NETworks (i.e.,~\OurMODEL) for 
antonym vs synonym distinction, that aim to capture and model the 
relation-specific properties of the antonyms and synonyms pairs in order to 
perform the classification task in a performance-enhanced manner.
Experimental evaluation using the benchmark datasets shows that \OurMODEL~
outperforms the existing research by a relative score of upto 1.8\% in F1-measure.
We release the codes for \OurMODEL~at
\url{https://github.com/asif6827/ICENET}.

\end{abstract}

\section{Introduction}

Antonyms vs synonyms distinction is a core challenge in natural 
language processing applications, including but not limited to:
sentiment analysis, machine translation, named entity typing etc.
Synonyms are defined as semantically related words, whereas antonyms 
are defined as semantically opposite words. For example ``disperse" 
and ``scatter" are synonyms, while ``disperse" and ``garner" are 
antonyms~\citep{2015_WET}.

Existing research on the antonym-synonym distinction is primarily 
categorized into pattern-based and embedding-based approaches.
Pattern-based approaches attempt to curate distinguishing 
lexico-syntactic patterns for the word pairs
~\citep{schwartz_2015, nguyen_2017}.
A major limitation of the pattern-based approaches is the 
sparsity of the feature space. Despite using massive 
data sets, the generalization attempts result in highly 
overlapping and noisy features, which further
deteriorate the model's performance.

Embedding based methods rely on the distributional hypothesis, i.e., 
\emph{``words that occur in the same contexts tend to have 
similar meanings"}~\citep{harris_1954}. These methods use 
widely available embedding resources to capture/compute 
the semantic relatedness of synonym and antonym pairs
~\citep{2016_dLCE, 2019_unravel}. 
~\citet{ali2019} proposed Distiller that uses non-linear 
projections to project the embedding vectors in task-specific 
dense sub-spaces.

The key challenge faced by existing embedding-based approaches 
is their inability to correctly model the inherent relation-specific 
properties among different relation pairs. These models 
mix different lexico-semantic relations and perform poorly 
when applied to a specific task~\citep{ali2019}.
Existing approaches, moreover, model each relation pair independently, 
which is not adequate for antonym and synonym relation pairs 
as these relation pairs exhibit unique properties that may 
be exploited by modeling the relation pair in correlation with other 
instances (discussed in detail in Section~\ref{proposed_approach}).

Keeping in view the above-mentioned challenges, in this paper, we propose
InterlaCed Encoder NETworks (\OurMODEL) for antonym vs synonym distinction. \OurMODEL~uses 
multiple different encoders to capture relation-specific properties of 
antonym and synonym pairs from pre-trained embeddings in order to 
augment the end-performance of the antonyms vs synonyms distinction task. 
Specifically, it uses:
(i) an encoder (ENC-1) to capture the symmetry of synonyms;
(ii) an encoder (ENC-2) to model the symmetry for antonyms; and
(iii) an encoder (ENC-3) to preserve the transitivity of the synonyms 
and trans-transitivity of antonym and synonym relation pairs by 
employing attentive graph convolutions. These relation-specific properties of 
antonym and synonym relation pairs are illustrated in Figure~\ref{fig:fig1}(a) and 
explained in Section~\ref{rel_explained}.


We are the first to make an attempt to use attentive graph convolutions for 
modeling the underlying characteristics of antonym and synonym relation pairs.
Note, this work is different from existing works using graph convolutional networks
for relational data, e.g.,~\cite{2018_RGCN}, as antonyms and synonyms possess 
unique properties which makes them different from relation pairs in the Knowledge 
Graphs (KG), e.g., FB15K, WN18~\cite{2013_TransE}.

\OurMODEL~is a shift from the existing instance-based modeling approaches to graph-based 
framing which allows effective information sharing across multiple instances at a time to perform the end 
classification in a performance-enhanced fashion. 
\OurMODEL~can be used with any available pre-trained embedding resources, 
which makes it more flexible than the existing approaches relying on huge 
text corpora. We summarize the major contributions of this paper as follows:

\begin{itemize}
\itemsep0em 
    \item We propose \OurMODEL, i.e., a combination of interlaced encoder networks to 
    refine relation-specific information from the pre-trained embeddings.
    \item \OurMODEL~is the first to use attentive graph convolutions for antonym vs 
    synonym distinction that provide a provision to analyze/classify a word pair 
    in correlation with multiple neighboring pairs/words, rather than independent 
    instant-level modeling.
    \item We demonstrate the effectiveness of the proposed model using benchmark 
    data sets. \OurMODEL~outperforms the existing models by a margin of upto 1.8\% 
    in terms of F1-measure.
\end{itemize}
\section{Related Work}

Earlier research on antonym synonym distinction attempts at capturing lexico-syntactic 
patterns between the word pairs co-occurring within the same sentence.

\citet{lin_2003} considered pharasal patterns: \emph{``from X to Y"},
and \emph{``either X or Y"} to identify synonyms amongst distributionally similar words.
\citet{baroni_2004} used co-occurrence statistics to discover synonyms and distinguish 
them from unrelated terms.
\citet{van_2006} used word alignment measures using parallel corpora from multiple 
different languages to capture synonyms.
\citet{lobanova_2010} used a set of seed pairs to capture patterns in the data and later 
used these patterns to extract new antonym pairs from text corpora.
\citet{roth_2014} proposed discourse markers as features alternate
to the lexico-syntactic patterns.
\citet{schwartz_2015} proposed automated routines to acquire a set of 
symmetric patterns for word similarity prediction.
\citet{nguyen_2017} proposed AntSynNET that uses a set of lexico-syntactic 
patterns between the word pairs within the same sentence captured over 
huge text corpora.

In the recent past, embedding models have received considerable research 
attention for antonyms vs synonyms distinction.
These models are based on distributional hypotheses, i.e., words 
with similar meanings co-occur in a similar context
~\citep{2014_word2vec,2014_glove, 2018_Fasttext}.
A major advantage offered by the embedding-based approaches is the freedom to 
curate and train embedding vectors for features extracted from text corpora. 
\citet{2014_adel} used skip-gram modeling to train embedding vectors using 
coreference chains. \citet{2016_dLCE} used lexical contrast information in 
the skip-gram model for antonym and synonym distinction.
\citet{2015_WET} uses dictionaries along with distributional information 
to detect probable antonyms.
\citet{ali2019} used a set of encoder functions to project the word embeddings 
in constrained sub-spaces in order to capture the relation-specific properties 
of the data. 
\citet{xie_2021} employed a mixture-of-experts framework based on a 
divide-and-conquer strategy. They used a number of localized experts
focused on different subspaces and a gating mechanism to formulate 
the expert mixture.

We observe some of the limitations of the existing work as follows.
The pattern-based approaches are limited owing to the noisy and overlapping 
nature of the patterns. 
The embedding models are limited by the challenges posed by the distributional 
nature of the word embeddings, e.g., in Glove embeddings top similar words for 
the word \emph{``small"} yields a combination of synonyms, antonyms, and irrelevant 
words~\cite{ali2019}.
\section{Background}

\subsection{Preliminaries}
\label{rel_explained}
Antonyms and synonyms are a special kind of relation pairs 
(denoted by $r_A$ and $r_S$) with unique properties, i.e., 
(a) antonyms possess symmetry, 
(b) synonyms exhibit symmetry and transitivity, 
(c) antonyms and synonyms when analyzed in combination demonstrate trans-transitivity.

These properties are depicted in Figure~\ref{fig:fig1} (a).
For ease of interpretation, we use $(h, r, t)$ to represent a 
relation tuple, where $h$ corresponds to the {``head"} and 
$t$ is the {``tail"} of relation $r$.
For word pair $(h, t)$ and relation $r$, symmetry implies 
$(h, r, t)$ iff $(t, r, h)$.
The transitivity between the relation implies: \emph{if $(h, r, t)$} 
and \emph{$(t, r, t^{'})$} hold then \emph{$(h, r, t^{'})$} also holds, 
as shown by the words {``nasty"} and {``horrible"} in Figure
~\ref{fig:fig1}(a).
Trans-transitivity implies: \emph{if $(h, r_A, t)$} and \emph{$(t, r_S, t^{'})$} hold 
then~\emph{$(h, r_A, t^{'})$} also holds, also illustrated between the 
words {``nasty"} and {``pleasing"}.

\subsection{KG Embeddings Methods}
\citet{ali2019} pointed out a key limitation of the translational embedding
methods (commonly used for KG embeddings) in modeling symmetric relations. For instance, for a symmetric relation $\rr$, 
it is not possible for translational embeddings to preserve both vector operations: 
$\hh + \rr = \tt$ and $\tt + \rr = \hh$ at the same time. This is also illustrated in
Figure~\ref{fig:fig1}(b), where we show $\tt^{'} \neq \tt$. 
For details refer to the original article by \citet{ali2019}.
Likewise, some of the key difference of our work from existing work, i.e., 
R-GCN by~\citet{2018_RGCN} are explained in Appendix~\ref{appndx_RGCN}.
\begin{figure}[h]
    \vspace{-1.7ex}
    \includegraphics[width=1.05\columnwidth]{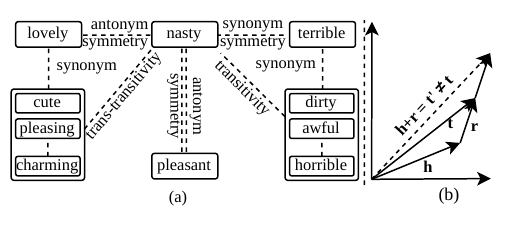} 
    \vspace{-5.7ex}
    \caption{(a) Properties of the antonym and synonym relation pairs, i.e., symmetry, 
    transitivity, and trans-transitivity; 
    (b) Limitation of translational embeddings in capturing the antonym and synonym relations~\cite{ali2019}.}
    \label{fig:fig1}
\end{figure}
\vspace{-2.7ex}

\eat{
Formally, using translational embeddings to model a symmetric relation, we have
\begin{align*}
\left.
\begin{aligned}
\hh + \rr - \tt = \Bepsilon_1 \quad\\
\hh + \rr - \tt = \Bepsilon_2 \quad
\end{aligned}
\right\}
\Rightarrow \quad
2\rr = \Bepsilon_1 + \Bepsilon_2 
\end{align*}
}

\eat{

These properties are depicted in Figure~\ref{fig:fig1} (a).
For ease of interpretation, we use $(h_1, r, t_1)$ to represent a 
relation pair, where $h_1$ corresponds to the \emph{``head"} and 
$t_1$ is the \emph{``tail"} of relation $r$.
For word pair $(h_1, t_1)$ and relation $r$, symmetry implies 
$(h_1, r, t_1)$ iff $(t_1, r, h_1)$.
The transitivity between the relation implies that 
\emph{if $(h_1, r, t_1)$ and $(t_1, r, t_2)$} holds then 
\emph{if $(h_1, r, t_2)$} also holds, as shown by the
words \emph{``nasty"} and \emph{``hostile"} in Figure
~\ref{fig:fig1}.
Trans-transitivity implies that 
\emph{if $(h_1, r_A, t_1)$ and $(t_1, r_S, t_2)$} holds then
\emph{$(h_1, r_A, t_2)$} also holds, illustrated between the 
words \emph{``nasty"} and \emph{``pleasing"} in Figure
~\ref{fig:fig1}. 

\begin{figure}[ht]
    \includegraphics[width=1.0\columnwidth]{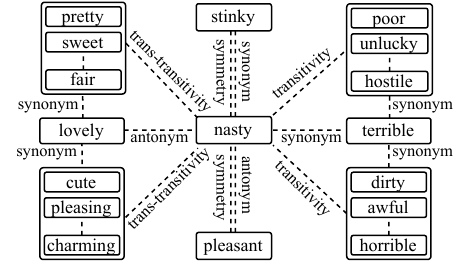} 
    \caption{Illustration of relation-specific properties of the antonym
    and synonym pairs, i.e., symmetry, transitivity, and trans-transitivity.}
    \label{fig:fig1}
\end{figure}
\vspace{-1.3ex}
}

\vspace{-0.7ex}
\section{Proposed Approach}
\label{proposed_approach}
\vspace{-0.7ex}
Given that existing KG embeddings are not able to model the relation-specific 
properties of the antonym and synonym pairs, we propose~\OurMODEL~that takes 
pre-trained word embeddings as inputs and projects them to low-dimensional space.
In order to ensure that low-dimensional space captures the relation-specific 
properties of the data to the best possible extent \OurMODEL~uses three different 
encoder networks. 
We call overall architecture as interlaced structure, because these networks are interconnected, 
i.e., (a) loss function of ENC-2 also depends upon ENC-1, 
(b) output of encoders (ENC-1, and ENC-2) is used as input to the ENC-3. 
Details about each encoder are as follows:

\vspace{-1.0ex}
\subsection{ENC-1}
\vspace{-1.0ex}
The goal of this encoder is to capture the symmetry of the synonym relation pairs. 
For this, we use a two-layered feed-forward function: 
$f_1(X) = \sigma W_{12} * \sigma(W_{11} * X + b_{11}) + b_{12}$ 
to project d-dimensional embeddings ($X \in \mathbf{R}^{d}$) to p-dimensions ($\mathbf{R}^{p}$).
Here $W_{11}$ and $W_{12}$ are the weight matrices; $b_{11}$ and $b_{12}$ are the bias terms.
For encoded word pairs to preserve symmetry among relation pairs, 
we employ negative sampling techniques. 
Specifically, we use a margin-based loss (shown in Equation~\ref{eq:SYN}) 
to project a word close to its true synonyms, while at the same time 
push it from irrelevant words. 
This formulation preserves the symmetry of the relation pair
owing to the commutative nature of the inner product. It is 
also justified by the fact: if $\xx_h$ is embedded close to $\xx_t$, 
then $\xx_t$ is also embedded close to $\xx_h$.
\vspace{-1.5ex}
\begin{equation}
  \label{eq:SYN}
    \begin{aligned}
        \text{L}_1 &= \sum_{(h, t) \in T_1}  \max(0, \gamma_1 -  \tanh(\inprod{f_1(\xx_h)}{f_1(\xx_t)})) \\
        & + \sum_{(h', t') \in T'_1} \max(0, \gamma_1 +  \tanh(\inprod{f_1(\xx_h^{'})}{f_1(\xx_t^{'})}))
    \end{aligned}
\end{equation}
Here $\gamma_1$ is the margin; $T_1$ corresponds to the synonym pairs; 
$\xx_{h}$, $\xx_{t}$ are the embedding vectors for head and tail words.
$T'_1$ is acquired by randomly replacing one of the words from 
the pairs in $T_1$ and/or using antonyms as negative samples.

\vspace{-1.0ex}
\subsection{ENC-2}
\vspace{-1.0ex}
This encoder aims to capture the symmetry for the antonym 
relation pairs. For this we use a two layered feed-forward function: 
$f_2(X) = \sigma W_{22} * \sigma(W_{21} * X + b_{21}) + b_{22}$ 
to project d-dimensional embeddings ($X \in \mathbf{R}^{d}$) to p-dimensions ($\mathbf{R}^{p}$).
Here $X \in \mathbf{R}^{d}$ corresponds to the pre-trained word embeddings; $W_{21}$ and 
$W_{22}$ are the weight matrices; $b_{21}$ and $b_{22}$ are the bias terms.
In order to preserve the symmetry of the antonym relations, we use another 
margin-based loss function (shown in Equation~\ref{eq:ANT}) to project a word 
close to its true antonyms, while at the same time push it from 
irrelevant words.
\vspace{-1.5ex}
\begin{equation}
  \label{eq:ANT}
  \begin{aligned} 
    \text{L}_2 &=  \sum_{(h, t) \in T_2} \max(0, \gamma_2 - \tanh(\inprod{f_2(\xx_h)}{f_1(\xx_t)})) \\
     & + \sum_{(h', t') \in T'_2} \max(0, \gamma_2 + \tanh(\inprod{f_2(\xx_h^{'})}{f_1(\xx_t^{'})}))
  \end{aligned}
\end{equation}
Note, for $\text{L}_2$ we use both functions, i.e., $f_1(X), f_2(X)$, that allows us to project 
$\xx_h$ close to its antonym $\xx_t$ as well as synonyms of $\xx_t$. Here again 
the symmetry of the relation is preserved by the commutative nature of the 
inner product.
$\gamma_2$ is the margin term, $T_2$ corresponds to the antonym pairs;
$\xx_{h}$, $\xx_{t}$ are the embedding vectors for head and tail words.
$T'_2$ is acquired by randomly replacing one of the words from 
the pairs in $T_2$ and/or using synonyms as negative samples.

Given that the encoders (ENC-1, ENC-2) use two different non-linear functions 
to project the pre-trained embeddings, it allows us to learn two projections for each word.
Later, we use all possible projection scores as indicators
for the word pair to be probable antonym and/or synonym pair.
This setting is different from the previous research that embeds 
synonyms close to each other, while antonyms are projected at an 
angle of 180$^{\circ}$~\cite{2015_WET} as it is hard to preserve 
the relation-specific properties for the resultant embeddings.

\eat{
\subsection{Overview}
\label{overview}
We aim to build a model to automatically classify a pair of 
words as antonyms and synonyms.
\OurMODEL~encompasses two different encoders and an attentive 
graph convolution layer to encode the information contained in 
the pre-trained embeddings in a way that it preserves the 
relational properties of the data to the best possible extent.
The encoders are responsible for preserving the symmetry of the 
relations, and the graph layer preserves the transitivity and 
trans-transitivity of the relation pairs. 
}

\subsection{ENC-3}
Finally, in order to preserve the transitivity of the synonym pairs and 
the trans-transitivity of antonym and synonym relation pairs in combination 
we propose an attentive graph convolutional encoder under transductive 
setting. We exploit the fact that the a word may be represented as a 
node in the graph, and each word may be surrounded by an arbitrary number of 
semantically related words as neighbouring nodes in the graph. We argue 
that this setting is more flexible in capturing the relation-specific 
properties involving arbitrary number of words, as it allows modeling the 
relation pairs in complete correlation with each other, which is more practical 
than modeling these pairs independent of each other. It also provides 
the provision for effective information sharing across the neighboring nodes
using attention weights. Similar ideas has already been applied to capture
the semantic-relatedness for embeddings trained for different 
languages~\cite{2023_GARI,2023_GRI}.

In our case, we use two different graphs, namely:
$\text{G}_h$, and $\text{G}_t$, for preserving the 
relations amongst the head and tail words respectively. 
We outline the graph construction process in Algorithm~\ref{alg:graphs}. 
It is explained as follows:
\begin{figure}[t]
\vspace{-0.7em}
\begin{algorithm}[H]
    \caption{Graph Construction}
    \label{alg:graphs}
    \textbf{Inputs:} {Embedding; $D=D_{tr}+D_{dev}+D_{test}$} 
    \textbf{Outputs:} {Graphs: $\text{G}_{h}, \text{G}_{t}$}
    \begin{algorithmic}[1]
        \State {$\{\text{Syn}_{h}, \text{Ant}_{h}\}_{h=1}^{V}$ $\gets$ {$\emptyset$};  $\text{G}_{t} \gets \emptyset$}
        \State {$\{\text{Syn}_{t}, \text{Ant}_{t}\}_{t=1}^{V}$ $\gets$ {$\emptyset$}; $\text{G}_{h} \gets \emptyset$}
        \State{Train $\text{M}_{init}(D_{tr};\text{L}_1,\text{L}_2)$}
        \For{$\text{inst}(h,t) \gets 1$ to $D$}
        \State{$y^{*}$ = $\text{score}(\text{M}_{init},\text{inst})$}
        \If{$y^{*} \geq \text{ANT}_{thr}$} 
            \State{$\text{Update} \{\text{Ant}_{h}; \text{Ant}_{t}$\}}
        \ElsIf{$y^{*} \leq \text{SYN}_{thr}$} 
            \State{$\text{Update} \{\text{Syn}_{h}; \text{Syn}_{t}$\}}
        \EndIf
        \EndFor

        \For{$pair \in \{\text{Syn}_{t}, \text{Ant}_{t}\}$}
        \State{$\text{G}_{h}$ $\gets$ $\text{G}_{h} \cup \{edge_{h}(pair)$\}}
        \EndFor
        \For{$pair \in \{\text{Syn}_{h}, \text{Ant}_{h}\}$}
        \State{$\text{G}_{t}$ $\gets$ $\text{G}_{t} \cup \{edge_{t}(pair)$\}}
        \EndFor
        \State \Return {$\text{G}_{h}$; $\text{G}_{t}$}    
    \end{algorithmic}
\end{algorithm}
\vspace{-3.1em}
\end{figure}

\paragraph{Graph Construction.}
The graph construction process uses data set $D$ and 300-d pre-trained 
Fasttext embeddings~\cite{2018_Fasttext} as inputs and returns two graphs 
$\text{G}_{h}$ and $\text{G}_{t}$ as output. The details are as follows.

Firstly, we initialize dictionaries \{$\text{Syn}_{h}$, $\text{Ant}_{h}$\} 
and \{$\text{Syn}_{t}$, $\text{Syn}_{t}$\} to store probable synonym 
and antonym pairs with head word $h$ and tail word $t$ respectively (lines 1-2).
We train a basic model ($\text{M}_{init}$) using the encoders (ENC-1 and ENC-2)
and available training data (line 3) in an end-to-end fashion. 
Later, $\text{M}_{init}$ is used to assign a score ($y^*$) to each pair 
in the data $D$ (line 6).
We use $y^*$ compared against the thresholds 
\{$\text{ANT}_{thr}$,$\text{SYN}_{thr}$,\} to update the data structures 
\{$\text{Ant}_h$, $\text{Ant}_t$\}, and \{$\text{Syn}_h$, $\text{Syn}_t$\} 
respectively (lines 6-9).
The core logic is: we add $\text{inst}(h, t)$ to $\text{Ant}_h$, if
(a) head word ($h$) corresponds to a key in $\text{Ant}_h$, 
(b) it is a probable antonym pair with ($y^{*} \geq \text{ANT}_{thr}$).
Later, we use the information in the dictionaries to construct 
the graphs (lines 12-16). 

We explain the construction of $\text{G}_{h}$ using the information 
in $\text{Syn}_t$, $\text{Ant}_t$, as follows. Given that $\text{Syn}_t$ 
contains the information about the list of probable synonym pairs with 
the tail word \emph{``t"}. In order to preserve the transitivity for the 
synonym pairs with tail \emph{``t"}, we formulate pairwise edges between 
the head terms in $\text{Syn}_t$. It is based on the assumption that head 
words of the relation pairs with the same tail, i.e., \emph{``t"} are 
likely to be synonyms of each other.
Likewise, $\text{Ant}_t$ contains the information about the list of 
probable antonym pairs with the tail word \emph{``t"}. In order to preserve 
the trans-transitivity of relation pairs with tail \emph{``t"}, we formulate 
pairwise edges between the head terms in $\text{Ant}_t$.
It is based on the assumption that the head words of the antonym relation 
pairs with same tail \emph{``t"} are likely to be synonyms of each other.
Eventually, we combine these edges to formulate the graph $\text{G}_{h}$.

We follow a similar procedure to construct the graph $\text{G}_{t}$ 
using information in $\text{Syn}_h$, $\text{Ant}_h$. Finally, we return
graphs $\text{G}_{h}$ and $\text{G}_{t}$ as the output of the 
graph construction process.

\vspace{-1ex}
\paragraph{Attentive Aggregation.}
The graph construction process surrounds each word in the graphs $G_{h}$ 
and $G_{t}$ by a set of probable synonyms. Later, it recomputes the 
representation of each word as an attentive aggregation of the neighbors.
For this, it uses the following layer-wise information propagation mechanism:
\begin{equation}
    \vspace{-0.7ex}
    L^{(i+1)} = \rho(\Tilde{\xi_{G}} L^{(i)} W_{i})
\end{equation}
where $\Tilde{\xi_{G}}= \bar{D}^{-1/2} (\xi_{G} + I) \bar{D}^{-1/2}$ is the
normalized symmetric marix, $\bar{D}$ is the degree matrix of $\xi_{G}$,
$\xi_{G}$ is the weighted adjacency matrix containing attention weights for $G$,
$L^{(i)}$ is the input representation from the previous layer, 
$W_{i}$ is the learn-able weight matrix.
We also add identity matrix $I$ to $\xi_{G}$ in order to allow 
self-connections for each word in the graphs. It allows the encoder
to analyze each word as a weighted combination of itself and its 
semantic neighbors. Our formulation for attentive graph convolutions is 
inspired by~\citet{2020_FGETRR}, and its non-euclidean variant~\citet{2021_FGETRH}.
Intuitive explanations in this regard are provided in Appendix~\ref{appendix_GCN}.

For \OurMODEL, we use a two-layered attentive graph convolution 
encoder with $\text{ReLU}$ non-linearity to generate the final 
representations of each word. Specifically, for the 
relation tuples $(h,r,t)$ in data $D$, the output of the encoders (ENC-1 and ENC-2) 
is separately processed by the attentive graph convolution networks 
to generate the final representations, as follows:
\begin{equation}         
  \label{eq:GCN}
      \begin{aligned}
      \XX_{hh} = \Tilde{\xi_{G_h}}(\text{ReLU} (\Tilde{\xi_{G_h}} \;f_1(X_{h}) W_{hh_1}) W_{hh_2} \\
      \XX_{ht} = \Tilde{\xi_{G_t}}(\text{ReLU} (\Tilde{\xi_{G_t}} \;f_1(X_{t}) W_{ht_1}) W_{ht_2} \\
      \XX_{th} = \Tilde{\xi_{G_h}}(\text{ReLU} (\Tilde{\xi_{G_h}} \;f_2(X_{h}) W_{th_1}) W_{th_2} \\
      \XX_{tt} = \Tilde{\xi_{G_t}}(\text{ReLU} (\Tilde{\xi_{G_t}} \;f_2(X_{t}) W_{tt_1}) W_{tt_2} 
      \end{aligned}
\end{equation}

Here $ f_1(X), f_2(X) \in \mathbf{R}^{p}$ are the outputs of the encoders (ENC-1, and ENC-2) used 
as inputs for ENC-3. $W_{i}$ are learn-able weights, $\XX_{i} \in \mathbf{R}^{q}$ are the outputs 
of attentive graph convolution. 
In order to train the attentive graph convolution network (ENC-3), 
we compute the score vectors \{$\xx_1, \xx_2$\} and \{$\xx_3, \xx_4$\} 
as indicative of synonymy and antonymy respectively.
\begin{equation}
  \label{eq:feature}
    \begin{aligned}
        \xx_1 = \cos{(\XX_{th}, \XX_{tt})}; & \; \xx_2 = \cos{(\XX_{hh}, \XX_{ht})} \\
        \xx_3 = \cos{(\XX_{hh}, \XX_{tt})}; & \; \xx_4 = \cos{(\XX_{ht}, \XX_{th})}
    \end{aligned}
\end{equation}
where $\cos(\XX,\YY)$ is the element-wise cosine of the vector 
pairs in $\XX$ and $\YY$.
We concatenate these scores to get the feature matrix: 
$X_\text{F}$ = [$\xx_1; \xx_2; \xx_3; \xx_4$], and use 
cross-entropy loss to train the encoder, shown in Equation~\ref{eq:CE}:
\begin{equation}
\label{eq:CE}
\centering
\resizebox{0.75\columnwidth}{!}{
    $\text{L}_3$ = $-\frac{1}{N}$ $\sum_{i=1}^{N} \log({p}(y_i|h_i, t_i))$}
\end{equation}
where ${p}(\yy \mid \xx_h, \xx_t) = softmax(\ww\xx_{F}+{\bb})$
with ${\hat{y}} = \argmax_y {p}(\yy|\xx_h, \xx_t)$, $\ww$ is 
the weight matrix and $\bb$ is the bias term.

\subsection{The Complete Model.}
Finally, we combine the loss functions of the individual encoder networks, i.e., 
$\text{L}_1 + \text{L}_2 + \text{L}_3$ as the loss function of \OurMODEL. 
We train the model in an end-to-end fashion.

\eat{
Note, this formulation also allows accumulating information from the 
higher-order neighborhood by stacking arbitrary numbers of layers.
$= \text{ENC}(\X) \in \mathbb{R}^{N\times{d}}$ is the input matrix that corresponds to the outputs of the encoder networks,
\textbf{Tell the difference
1. Not computed for each pair.
2. We use hard attention weights, based on the degree of their similarity.
}
\eat{
\begin{equation}
\label{eq:layers}
\centering
    \resizebox{0.88\columnwidth}{!}{
    X_m = \widetilde{(\xi_{ij} \odot A)}(ReLU(\widetilde{(\xi_{ij} \odot A)}X W_0)) W_1}
\end{equation}}
}
\section{Experiments and Results}

\subsection{Datasets}
We evaluate the proposed approach on two different data sets: 
(a) A benchmark data set by~\cite{nguyen_2017} manually curated 
from WordNet and Wordnik\footnote{http://www.wordnik.com}. 
It encompasses randomly split synonyms and antonym pairs 
corresponding to three word classes (adjective, noun, and verb). 
(b) A lexical split curated by~\cite{xie_2021}.
For both data sets, the ratio between the antonyms and synonym 
pairs within each word class is approximately 1:1. The 
statistics of each data set are shown in Table~\ref{tab:data}.
\eat{Note, owing to very limited availability of multi-sense data sets 
for the antonym vs synonym distinction task, current formulation 
of \OurMODEL~is not evaluated under multi-sense settings.}

\begin{table}[t]
    \centering
    \resizebox{1.02\columnwidth}{!}{
\begin{tabular}{c|ccc|ccc}
\hline
\multirow{2}{*}{word class} & \multicolumn{3}{c}{(a) Random} & \multicolumn{3}{|c}{(b) Lexical} \\
                            & train     & dev     & test     & train        & dev       & test    \\
\hline
Adjective                   & 5562      & 398     & 1986     & 4227         & 303       & 1498    \\
Noun                        & 2836      & 206     & 1020     & 2667         & 191       & 954     \\
Verb                        & 2534      & 182     & 908      & 2034         & 146       & 712     \\          
\hline
\end{tabular}}
\vspace{-1.7ex}
    \caption{Antonym/Synonym distinction datasets}
        \vspace{-2.7ex}
    \label{tab:data}
\end{table}

\subsection{Experimental Settings}
Similar to the baseline methods, for main experimentation we report the results 
using random split and 300-d Fasttext embeddings~\citep{2018_Fasttext} trained 
on wiki-news corpus. Results using the dLCE embeddings and lexical split of the data 
are discussed in Section~\ref{discussion}. The embedding vectors for the OOV 
tokens are randomly initialized. For model training, we use Adam optimizer~\cite{2014_adam} 
with learning rate=0.001. The values for $\text{SYN}_{thr}$ and $\text{ANT}_{thr}$ 
are set to 0.15 and 0.10 respectively. 
For $\text{L}_1$ and $\text{L}_2$ the values for the margin terms are: $\gamma_1 = \gamma_2 = 0.9$. 
Output dimensionality of ENC-1 and ENC-2 is 80d and for ENC-3 is 60d.
We used TensorFlow toolkit (version 2.12) to run the experiments.
We report mean and standard deviation of the scores computed over five runs of the experiments. 
All experiments were performed using Intel Core-i9-10900X CPU, and Nvidia 3090Ti GPU. 
On this GPU, a single run of the experiments takes approximately thirty minutes.

\begin{table*}[t]
\centering
\resizebox{2.10\columnwidth}{!}{%
\begin{tabular}{lll|ccc|ccc|ccc}
\hline
\multicolumn{3}{l|}{\multirow{2}{*}{Methodology}} & \multicolumn{3}{c|}{Adjective} & \multicolumn{3}{c|}{Verb} & \multicolumn{3}{c}{Noun} \\
\multicolumn{3}{c|}{}                               &   P  &   R      & F1      & P      & R      & F1     & P      & R      & F1     \\
\hline
\multicolumn{3}{l|}{Baseline-1 (Random vectors)}  & 0.657  & 0.665  & 0.661  & 0.782  & 0.819  & 0.800  & 0.783  & 0.751  & 0.767 \\
\multicolumn{3}{l|}{Baseline-2 (w/o Graph conv.)} & 0.828  & 0.909  & 0.867    & 0.837  & 0.915  & 0.879  & 0.818  & 0.818  & 0.818 \\
\multicolumn{3}{l|}{AntSynNet~\cite{nguyen_2017}} & 0.750  & 0.798  & 0.773  & 0.717  & 0.826  & 0.768  & 0.807  & 0.827  & 0.817  \\
\multicolumn{3}{l|}{Parasiam~\cite{2019_unravel}} & 0.855  & 0.857  & 0.856  & 0.864  & \underline{0.921}  & 0.891  & 0.837  & 0.859  & 0.848  \\
\multicolumn{3}{l|}{Distiller~\cite{ali2019}}     & 0.854  & \underline{0.917}    & 0.884 & 0.871  & 0.912  & 0.891  & 0.823  & \underline{0.866}  & 0.844  \\
\multicolumn{3}{l|}{MoE-ASD~\cite{xie_2021}}      & \underline{0.878}    & 0.907    & \underline{0.892}   & \underline{0.895}  & 0.920   & \underline{0.908}  & \underline{0.841}  & \textbf{0.900} & \underline{0.869}  \\
\hline
\multicolumn{3}{l|}{\OurMODEL}       & \textbf{0.896}$\pm$ 0.0005 & \textbf{0.919}$\pm$ 0.0005 & \textbf{0.908}$\pm$ 0.0005 & \textbf{0.899}$\pm$ 0.001 & \textbf{0.932}$\pm$ 0.001& \textbf{0.915}$\pm$ 0.001 & \textbf{0.895}$\pm$ 0.001  & 0.871$\pm$ 0.001  & \textbf{0.883}$\pm$ 0.001  \\
\hline
\end{tabular}}
\vspace{-1.7ex}
\caption{\OurMODEL~performance comparison using random split}
\vspace{-2.7ex}
\label{tab:res1}
\end{table*}

\subsection{Baseline Models}
In order to test the effectiveness of \OurMODEL, we design 
two baseline models. 
Baseline-1 aims to analyze the ability of \OurMODEL~to encode 
the information in the pre-trained embeddings. For this, we use  
random vectors in place of pre-trained embeddings.
Baseline-2 aims to analyze the ability of graph convolutions 
to preserve relation-specific properties. For this, we use 
a basic variant of \OurMODEL~relying only on the ENC-1 
and ENC-2.

We also compare \OurMODEL~with
existing state-of-the-art research on the antonym-synonym distinction task, 
i.e., (i) AntSynNET by~\citet{nguyen_2017},
(ii) Parasiam by~\citet{2019_unravel}, (iii) Distiller by~\citet{ali2019}, 
and (iv) MoE-ASD by~\citet{xie_2021}.
For all these models, we report the scores reported in the original 
papers, as they are computed using the same data settings as that of ours.

\subsection{Main Results}
\label{sec:main_res}
The performance comparison of \OurMODEL~is reported in 
Table~\ref{tab:res1}. 
For these results, we use the random split of the data and 300-d 
Fasttext embeddings. We boldface overall best scores with 
previous state-of-the-art underlined. A low variance of the results shows that 
\OurMODEL~yields a stable performance across multiple runs.

Comparing the performance of \OurMODEL~against the previous 
state-of-the-art, we observe, for the adjective data sets, 
the~\OurMODEL~ outperforms existing best by 2.1\%, 0.2\% and 
1.8\% for precision, recall, and F1 scores respectively.
For the verbs data set, it outweighs the precision, recall and 
F1 score by 0.45\%, 1.19\%, and 0.77\% respectively.
For the nouns data set the improvement in performance for the
precision and F1-scores is 6.42\%, and 1.61\%.

Analyzing the performance of \OurMODEL~against the baseline 
models, a significant decline in the performance for the baseline-1 
shows that pre-trained embeddings carry a significant amount of 
relation-specific information which is refined by 
\OurMODEL~in a performance-enhanced fashion.
Likewise, the performance comparison against the baseline-2 
shows that attentive graph convolutions help the \OurMODEL~in capturing 
probable relation pairs by using the relation-specific properties, i.e.,
symmetry, transitivity, and trans-transitivity to the best possible 
extent, which in turn boosts the end performance of the model.

These results show the impact of using attentive graph convolutions
for the distinction task. It affirms our hypothesis that graph 
convolutions offer an optimal setting to model the relation-specific 
data because it provides the provision for information sharing across 
semantically related words, rather than modeling data instances 
completely independently of each other.
\vspace{-1.3ex}
\section{Analyses}
\label{discussion}
\vspace{-1.3ex}
In this section, we perform a detailed analyses of 
the~\OurMODEL~under different settings, 
namely: (i) dLCE embeddings~\cite{2016_dLCE}, (ii) Lexical split,
(iii) Ablation analysis, and (iv) Error analyses.

\begin{table*}[t]
\centering
\resizebox{2.10\columnwidth}{!}{%
\begin{tabular}{lll|ccc|ccc|ccc}
\hline
\multicolumn{3}{l|}{\multirow{2}{*}{Methodology}}   & \multicolumn{3}{c|}{Adjective} & \multicolumn{3}{c|}{Verb} & \multicolumn{3}{c}{Noun} \\
\multicolumn{3}{c|}{}                               &   P      &   R      & F1      & P      & R      & F1     & P      & R      & F1     \\
\hline
\multicolumn{3}{l|}{AntSynNet~\cite{nguyen_2017}}    & 0.763    & 0.807    & 0.784   & 0.743  & 0.815  & 0.777  & 0.816  & 0.898  & 0.855  \\
\multicolumn{3}{l|}{Parasiam~\cite{2019_unravel}}   & 0.874    & \textbf{0.950}    & 0.910   & 0.837  & \textbf{0.953}  & 0.891  & 0.847  & 0.939  & 0.891  \\
\multicolumn{3}{l|}{Distiller~\cite{ali2019}}       & 0.912    & 0.944    & 0.928   & 0.899  & 0.944  & 0.921  & 0.905  & 0.918  & 0.911  \\
\multicolumn{3}{l|}{MoE-ASD~\cite{xie_2021}}        & 0.935    & 0.941    & 0.938   & \textbf{0.914} & 0.944  & 0.929  & 0.920  & 0.950  & 0.935  \\
\hline
\multicolumn{3}{l|}{\OurMODEL}          & \textbf{0.936}$\pm${0.0002}  & {0.945}$\pm${0.0002} & \textbf{0.940}$\pm${0.0002} & 0.913$\pm${0.001}  & \textbf{0.953}$\pm${0.001}  & \textbf{0.933}$\pm${0.001}  & \textbf{0.925}$\pm${0.001}  & \textbf{0.953}$\pm${0.001}  & \textbf{0.939}$\pm${0.001}  \\
\hline
\end{tabular}
}
\vspace{-1.7ex}
\caption{\OurMODEL~performance comparison using random split and dLCE Embeddings}
\vspace{-2.7ex}
\label{tab:res2}
\end{table*}

\vspace{-1ex}
\subsection{dLCE Embeddings}
\vspace{-1ex}
Results for \OurMODEL~ using random split and dLCE embeddings 
are shown in Table~\ref{tab:res2}. We also report the scores 
for the previous research using the same test settings (i.e., 
data split and embeddings).
These results show \OurMODEL~outperforms the existing research
yielding a higher value of F1-score across all three data 
categories (adjective, verb and noun).
These results compared to the results in Table~\ref{tab:res1}
(using fasttext embeddings) show that dLCE embeddings being trained 
on lexical contrast information carry more distinctive information 
for the distinction task compared to generalized pre-trained word embeddings.

\vspace{-1ex}
\subsection{Lexical Split}
\vspace{-1ex}
In this subsection, we analyze the results of \OurMODEL~ 
corresponding to the lexical split of the antonym synonym 
distinction task~\cite{xie_2021}. Note, the lexical split 
assumes no overlap across train, dev, and test splits 
in order to avoid lexical memorization~\cite{2016_shwartz}.
Generally, the lexical split is considered a much tough 
evaluation setting compared to the random split, as it 
doesn't allow information sharing across different data 
splits based on overlapping vocabulary.

For the lexical split, the results for both dLCE embeddings 
and Fasttext embeddings are shown in Table ~\ref{tab:res3}.
Comparing the performance of our model against existing 
research, it is evident for both embeddings, i.e., Fasttext and dLCE,
~\OurMODEL~yields a higher F1 measure compared to the existing models.
\begin{table*}[t]
\centering
\resizebox{2.10\columnwidth}{!}{%
\begin{tabular}{lll|ccc|ccc|ccc}
\hline
\multicolumn{3}{l|}{\multirow{2}{*}{Adjacency}} & \multicolumn{3}{c|}{Adjective} & \multicolumn{3}{c|}{Verb} & \multicolumn{3}{c}{Noun} \\
\multicolumn{3}{c|}{} &   P      &   R      & F1      & P      & R      & F1     & P      & R      & F1     \\
\hline
\multicolumn{3}{l|}{\OurMODEL~(A1 = Random)}                     & 0.862  & 0.863 & 0.863 & 0.799 & 0.894 & 0.844 & 0.816  & 0.863 & 0.839  \\
\multicolumn{3}{l|}{\OurMODEL~(A2 = Identity)}                   & 0.849  & 0.886 & 0.867 & 0.761 & 0.896 & 0.823 & 0.830  & 0.861 & 0.845  \\
\multicolumn{3}{l|}{\OurMODEL~(A3 = Fasttext)}                   & 0.880  & 0.873 & 0.877 & 0.867 & 0.930 & 0.897 & 0.851 & \textbf{0.873}  & 0.862  \\
\multicolumn{3}{l|}{\OurMODEL~(A4 = $\text{M}_{init}$)}          & 0.881  & 0.909 & 0.895 & \textbf{0.899} & 0.925 & 0.912 & 0.874  & 0.867 & 0.870 \\
\multicolumn{3}{l|}{\OurMODEL~(A5 = Weighted-$\text{M}_{init}$)} & \textbf{0.896}$\pm${0.0005}  & \textbf{0.919}$\pm${0.0005} & \textbf{0.908}$\pm${0.0005} & 0.898$\pm${0.001} & \textbf{0.932}$\pm${0.001} & \textbf{0.915}$\pm${0.001} & \textbf{0.895}$\pm${0.001} & 0.871$\pm${0.001} & \textbf{0.883}$\pm${0.001}  \\
\hline
\end{tabular}}
\vspace{-1.7ex}
\caption{\OurMODEL~performance comparison using different adjacency matrices and random data split}
\vspace{-1.7ex}
\label{tab:abl1}
\end{table*}

\vspace{-1.7ex}
\subsection{Ablation Analyses}
\vspace{-1ex}

The core focus of \OurMODEL~is to employ attentive graph convolutions in order to capture the 
relation-specific properties of antonym and synonym pairs in order to perform the distinction 
task in a robust way. In order to simplify things, we deliberately don't include any 
hand-crafted features, e.g., negation prefixes etc., as a part of~\OurMODEL. 

For the ablation analyses of \OurMODEL, we: 
(a) compare the performance of~\OurMODEL~with and without attentive graph convolutions, 
(b) analyze the impact of different attention weights. 
\paragraph{(a) Impact of attentive convolutions.}
In order to analyze the impact of attentive graph convolutions,
we train a variant of \OurMODEL~ encompassing only the encoder
networks. Note, we also used a similar model in Section~\ref{sec:main_res} 
(shown as baseline-2 in Table~\ref{tab:res1}), however, the end 
goal of this analysis is to dig out a few example pairs 
that benefited especially from the attentive graph convolutions.

Some of the synonym and antonym word pairs that were corrected 
by attentive convolutions include: \{(lecture, reprimand), (single,retire)\} 
and \{(tender,demand), (file, rank)\} respectively. These word pairs 
were not easy to categorize otherwise by the variant of
\OurMODEL~without graph convolutions.
This shows the significance of the attentive convolutions in 
acquiring relation-specific information from semantically related 
neighbors that was helpful to reinforce the classification decision.
\begin{table*}[t]
\centering
\resizebox{2.10\columnwidth}{!}{
\begin{tabular}{l|l|ccc|ccc|ccc}
\hline
\multirow{2}{*}{Embedding} & \multirow{2}{*}{Model} & \multicolumn{3}{c|}{Adjective} & \multicolumn{3}{c|}{Verb} & \multicolumn{3}{c}{Noun} \\
                               &                        & P        & R        & F1      & P      & R      & F1     & P      & R      & F1     \\
                               \hline
\multirow{3}{*}{FastText}      & Parasiam~\cite{2019_unravel} & 0.694    & 0.866    & 0.769   & 0.642  & \textbf{0.824}  & 0.719  & 0.740   & 0.759  & 0.748  \\
                               & MoE-ASD~\cite{xie_2021}      & 0.808    & 0.810     & 0.809   & \textbf{0.830}   & 0.693  & 0.753 & \textbf{0.846}  & 0.722  & 0.776  \\
                               & \OurMODEL                    & 0.760$\pm${0.0005}    & \textbf{0.870}$\pm${0.0005}    & \textbf{0.815}$\pm${0.0005}    & 0.740$\pm${0.001}      & 0.777$\pm${0.001}    & \textbf{0.758}$\pm${0.001}     & 0.763$\pm${0.002} & \textbf{0.826}$\pm${0.002} & \textbf{0.793}$\pm${0.002}      \\
                               \hline
\multirow{3}{*}{dLCE}          & Parasiam~\cite{2019_unravel} & 0.768    & 0.952    & 0.850    & 0.769  & 0.877  & 0.819  & 0.843  & 0.914  & 0.876  \\
                               & MoE-ASD~\cite{xie_2021} & \textbf{0.877}    & 0.908    & 0.892   & \textbf{0.860}  & 0.835  & 0.847  & \textbf{0.912}  & 0.869  & 0.890  \\
                               & \OurMODEL               & 0.835$\pm${0.0004}   & \textbf{0.971}$\pm${0.0004}  & \textbf{0.898}$\pm${0.0004}   & 0.793$\pm${0.002}  & \textbf{0.938}$\pm${0.002}  & \textbf{0.859}$\pm${0.002}  & 0.886$\pm${0.001}  & \textbf{0.915}$\pm${0.001}  & \textbf{0.900}$\pm${0.001}      \\
                               \hline
\end{tabular}}
\vspace{-1.7ex}
\caption{Antonym/Synonym distinction performance for the lexical split}
\vspace{-2.7ex}
\label{tab:res3}
\end{table*}

\paragraph{(b) Varying attention weights.}
We also analyze the impact of different attention weights 
on the end performance of the model. Corresponding results
are shown in Table~\ref{tab:abl1}. For these experiments, 
we use five different types of attention weights, yielding 
adjacency matrices: A1, A2, A3, A4, and A5 in Table~\ref{tab:abl1}. 
We use hard attention weights that are not fine-tuned during 
the model training.
The graphs ($\text{G}_{h}$ and $\text{G}_{t}$) used in these experiments 
correspond to the best performing variant of \OurMODEL.

For A1, we use random values as attention weights, i.e., 
we randomly assign a value to each word pair from the 
range (0.1 $\sim$ 0.9). 
For A2, we use the identity as the adjacency matrix for 
the word pairs in the graphs, i.e., we completely ignore 
the effect of graph convolutions. For A3, we use the 
embedding similarity scores of the fasttext embeddings 
as the attention scores. This setting is based on the 
distributional hypothesis, i.e., distributionally similar 
words get higher scores.
For A4, we use the embedding similarity scores from the output 
of  ENC-1 network for the model 
$\text{M}_{init}$, trained entirely using two encoder 
networks. The motivation for using these scores as attention 
weights is the fact that ENC-1 is responsible for capturing 
the synonym pairs, so it will assign a higher score to probable 
synonyms, and a relatively lower score to probable antonyms.

For A5, we use attention weights similar to the setting of 
A4 with the difference that we down-scale the weights 
for probably erroneous edges in the graph. For less confident 
relation pairs with scores closer to the thresholds, i.e., 
$\text{ANT}_{thr}$, $\text{SYN}_{thr}$, we simply downscale the 
attention weight by half. This setting in turn limits the error propagation in 
the end-model caused by the erroneous edges in the graphs.

Results in Table~\ref{tab:abl1} show that \OurMODEL~(A5),
outperforms other variants of attention weights.
A similar performance is observed by the model ~\OurMODEL~(A4).
Relatively lower scores for the models using the random values and 
identity matrices as attention weights show the significance of sharing information amongst 
semantically related neighbors in an appropriate proportion in order to 
perform the end task in a performance-enhanced way.
Likewise, the score for \OurMODEL~(A3) show that by default 
the distributional scores of the pre-trained embeddings are not 
suitable for the end task.
These analyses clearly indicate that the choice of attention 
weight plays a vital role in capturing the properties of the 
data.

\subsection{Error Analyses}
\vspace{-0.7ex}

For the variant of~\OurMODEL~using random split and Fasttext embeddings, 
we collect a sample of approximately fifty error cases for each word 
class (adjectives, verbs, and nouns) to analyze the most probable reasons 
for the errors. 
We broadly categorize the errors into the following different categories: 
(a) the inability of input embeddings to cater to multiple senses, 
(b) the distributional embeddings for out-of-vocabulary (OOV) and/or rare words, and 
(c) other cases, e.g., negation prefix, errors with unknown reasons etc.

We separately report the number of erroneous edges/neighbours in the 
graphs: $\text{G}_{h}$ and $\text{G}_{t}$. Information propagation over these 
erroneous edges may also lead to the classification errors, however, it is 
hard to quantify such errors.

For adjectives, almost 25\% errors correspond to the sense category,
20\% errors are caused by rare words and/or OOV tokens, 
and the rest errors are attributed to negation prefixes and 
unknown reasons.
For nouns, 30\% errors belong to the sense category,
12.5\% errors result due to rare words and/or OOV tokens,
with the rest of the errors assigned to the negation prefixes and 
unknown reasons.
For verbs, 13\% errors correspond to the sense category,
15\% errors are caused by rare words/OOV tokens and the rest of 
the errors may be attributed to negation prefixes and unexplained reasons.
Regarding erroneous neighborhoods in the graphs, almost 11\%, 
12\% and 5\% neighbors of the graphs for adjectives, 
nouns and verbs respectively are erroneous, which deteriorate the 
end-performance of~\OurMODEL~by error propagation through 
attentive convolutions.

Considering the impact of different error categories on the end 
performance of~\OurMODEL. For multi-sense tokens the distributional 
embedding vectors are primarily oriented in the direction of the 
most prevalent sense of the underlying  training corpora, which may 
be different from the sense in the word pair resulting in 
misclassifications. 
For example, ``clean" and ``blue" are two synonym words in 
the adjective dataset. Looking at the most similar words in the 
fasttext embeddings, we can see that the embedding vector for 
the word ``blue" is more related to the colors, which makes 
it sense-wise different from the word ``clean" which is 
more related to cleanliness.
If we use these words to explain the properties of water, 
then these words are synonyms, however, it is not evident 
unless we explicitly consider the context along with word pair.
Note, the phenomenon of multiple senses of a given word is more dominant 
among nouns compared with that of verbs and adjectives. This is also 
evident by a relatively lower performance of nouns relative to other 
word classes.
For rare and OOV words, the embedding vectors are not adequately 
trained and their role in the end model is no better than the 
random vectors. This in turn limits the encoder networks of 
\OurMODEL~to encode relation-specific information.

\eat{
\OurMODEL~completely relies on the initial model
$\text{M}_{init}$ for the construction of 
graphs: $\text{G}_{h}$ and 
$\text{G}_{t}$.
}
\vspace{-1.7ex}
\section{Conclusion \& Future Work}
\vspace{-1.7ex}
In this work we propose~\OurMODEL, which uses a set of interlaced
encoder networks to capture the relation-specific properties of 
antonym and synonym pairs, i.e., symmetry, transitivity, and trans-transitivity, 
in order to perform antonyms vs synonyms distinction task. 
Results show that \OurMODEL~outperforms the existing research by a relative 
score of up to 1.8\% for F1-measure. Some promising future directions include:
(i) using domain-specific text corpora along with training seeds,
(ii) strategy to cut down the attention weights for the erroneous edges.

\section{Limitations}
Some of the core limitations of the \OurMODEL~ are as follows:
\begin{enumerate}
\itemsep0em 
    \item Nouns and adjectives exhibit multiple different senses, which requires the need for 
    the contextual information along with the word pair in order to model them. However, owing 
    to unavailability of multi-sense data sets for the antonym vs synonym distinction task, 
    current formulation of \OurMODEL~does not support multi-sense settings.
    \item Erroneous edges in the adjacency graphs produced by $M_{init}$ lead to error propagation. 
    There is a need for an appropriate attention mechanism based on the semantics of the data.
    \item The embeddings corresponding to the rare words and OOV tokens need to be initialized 
    as a weighted average of semantically related tokens rather than random initialization.
\end{enumerate}

\bibliography{custom}

\begin{thebibliography}{27}
\expandafter\ifx\csname natexlab\endcsname\relax\def\natexlab#1{#1}\fi

\bibitem[{Adel and Sch{\"u}tze(2014)}]{2014_adel}
Heike Adel and Hinrich Sch{\"u}tze. 2014.
\newblock Using mined coreference chains as a resource for a semantic task.
\newblock In \emph{Proceedings of the 2014 conference on empirical methods in
  natural language processing (EMNLP)}, pages 1447--1452.

\bibitem[{Ali et~al.(2023{\natexlab{a}})Ali, Alshmrani, Qin, Hu, and
  Wang}]{2023_GARI}
Muhammad Ali, Maha Alshmrani, Jianbin Qin, Yan Hu, and Di~Wang.
  2023{\natexlab{a}}.
\newblock Gari: Graph attention for relative isomorphism of arabic word
  embeddings.
\newblock In \emph{Proceedings of ArabicNLP 2023}, pages 181--190.

\bibitem[{Ali et~al.(2023{\natexlab{b}})Ali, Hu, Qin, and Wang}]{2023_GRI}
Muhammad Ali, Yan Hu, Jianbin Qin, and Di~Wang. 2023{\natexlab{b}}.
\newblock Gri: Graph-based relative isomorphism of word embedding spaces.
\newblock In \emph{Findings of the Association for Computational Linguistics:
  EMNLP 2023}, pages 11304--11313.

\bibitem[{Ali et~al.(2020)Ali, Sun, Li, and Wang}]{2020_FGETRR}
Muhammad~Asif Ali, Yifang Sun, Bing Li, and Wei Wang. 2020.
\newblock Fine-grained named entity typing over distantly supervised data based
  on refined representations.
\newblock In \emph{Proceedings of the AAAI Conference on Artificial
  Intelligence}, volume~34, pages 7391--7398.

\bibitem[{Ali et~al.(2021)Ali, Sun, Li, and Wang}]{2021_FGETRH}
Muhammad~Asif Ali, Yifang Sun, Bing Li, and Wei Wang. 2021.
\newblock Fine-grained named entity typing over distantly supervised data via
  refinement in hyperbolic space.
\newblock \emph{arXiv preprint arXiv:2101.11212}.

\bibitem[{Ali et~al.(2019)Ali, Sun, Zhou, Wang, and Zhao}]{ali2019}
Muhammad~Asif Ali, Yifang Sun, Xiaoling Zhou, Wei Wang, and Xiang Zhao. 2019.
\newblock Antonym-synonym classification based on new sub-space embeddings.
\newblock In \emph{Proceedings of the AAAI Conference on Artificial
  Intelligence}, volume~33, pages 6204--6211.

\bibitem[{Baroni and Bisi(2004)}]{baroni_2004}
Marco Baroni and Sabrina Bisi. 2004.
\newblock Using cooccurrence statistics and the web to discover synonyms in a
  technical language.
\newblock In \emph{LREC}.

\bibitem[{Bordes et~al.(2013)Bordes, Usunier, Garcia-Duran, Weston, and
  Yakhnenko}]{2013_TransE}
Antoine Bordes, Nicolas Usunier, Alberto Garcia-Duran, Jason Weston, and Oksana
  Yakhnenko. 2013.
\newblock Translating embeddings for modeling multi-relational data.
\newblock \emph{Advances in neural information processing systems}, 26.

\bibitem[{Etcheverry and Wonsever(2019)}]{2019_unravel}
Mathias Etcheverry and Dina Wonsever. 2019.
\newblock Unraveling antonym’s word vectors through a siamese-like network.
\newblock In \emph{Proceedings of the 57th Annual Meeting of the Association
  for Computational Linguistics}, pages 3297--3307.

\bibitem[{Goldberg and Levy(2014)}]{2014_word2vec}
Yoav Goldberg and Omer Levy. 2014.
\newblock word2vec explained: deriving mikolov et al.'s negative-sampling
  word-embedding method.
\newblock \emph{arXiv preprint arXiv:1402.3722}.

\bibitem[{Grave et~al.(2018)Grave, Bojanowski, Gupta, Joulin, and
  Mikolov}]{2018_Fasttext}
Edouard Grave, Piotr Bojanowski, Prakhar Gupta, Armand Joulin, and Tomas
  Mikolov. 2018.
\newblock Learning word vectors for 157 languages.
\newblock In \emph{Proceedings of the International Conference on Language
  Resources and Evaluation (LREC 2018)}.

\bibitem[{Harris(1954)}]{harris_1954}
Zellig~S Harris. 1954.
\newblock Distributional structure.
\newblock \emph{Word}, 10(2-3):146--162.

\bibitem[{Kingma and Ba(2014)}]{2014_adam}
Diederik~P Kingma and Jimmy Ba. 2014.
\newblock Adam: A method for stochastic optimization.
\newblock \emph{arXiv preprint arXiv:1412.6980}.

\bibitem[{Kipf and Welling(2016)}]{2016_GCN}
Thomas~N Kipf and Max Welling. 2016.
\newblock Semi-supervised classification with graph convolutional networks.
\newblock \emph{arXiv preprint arXiv:1609.02907}.

\bibitem[{Lin et~al.(2003)Lin, Zhao, Qin, and Zhou}]{lin_2003}
Dekang Lin, Shaojun Zhao, Lijuan Qin, and Ming Zhou. 2003.
\newblock Identifying synonyms among distributionally similar words.
\newblock In \emph{IJCAI}, volume~3, pages 1492--1493.

\bibitem[{Lobanova et~al.(2010)Lobanova, Van~der Kleij, and
  Spenader}]{lobanova_2010}
Anna Lobanova, Tom Van~der Kleij, and Jennifer Spenader. 2010.
\newblock Defining antonymy: A corpus-based study of opposites by
  lexico-syntactic patterns.
\newblock \emph{International Journal of Lexicography}, 23(1):19--53.

\bibitem[{Nguyen et~al.(2016)Nguyen, Walde, and Vu}]{2016_dLCE}
Kim~Anh Nguyen, Sabine Schulte~im Walde, and Ngoc~Thang Vu. 2016.
\newblock Integrating distributional lexical contrast into word embeddings for
  antonym-synonym distinction.
\newblock \emph{arXiv preprint arXiv:1605.07766}.

\bibitem[{Nguyen et~al.(2017)Nguyen, Walde, and Vu}]{nguyen_2017}
Kim~Anh Nguyen, Sabine Schulte~im Walde, and Ngoc~Thang Vu. 2017.
\newblock Distinguishing antonyms and synonyms in a pattern-based neural
  network.
\newblock \emph{arXiv preprint arXiv:1701.02962}.

\bibitem[{Ono et~al.(2015)Ono, Miwa, and Sasaki}]{2015_WET}
Masataka Ono, Makoto Miwa, and Yutaka Sasaki. 2015.
\newblock Word embedding-based antonym detection using thesauri and
  distributional information.
\newblock In \emph{Proceedings of the 2015 Conference of the North American
  Chapter of the Association for Computational Linguistics: Human Language
  Technologies}, pages 984--989.

\bibitem[{Pennington et~al.(2014)Pennington, Socher, and Manning}]{2014_glove}
Jeffrey Pennington, Richard Socher, and Christopher~D Manning. 2014.
\newblock Glove: Global vectors for word representation.
\newblock In \emph{Proceedings of the 2014 conference on empirical methods in
  natural language processing (EMNLP)}, pages 1532--1543.

\bibitem[{Roth and Im~Walde(2014)}]{roth_2014}
Michael Roth and Sabine~Schulte Im~Walde. 2014.
\newblock Combining word patterns and discourse markers for paradigmatic
  relation classification.
\newblock In \emph{Proceedings of the 52nd Annual Meeting of the Association
  for Computational Linguistics (Volume 2: Short Papers)}, pages 524--530.

\bibitem[{Schlichtkrull et~al.(2018)Schlichtkrull, Kipf, Bloem, Van Den~Berg,
  Titov, and Welling}]{2018_RGCN}
Michael Schlichtkrull, Thomas~N Kipf, Peter Bloem, Rianne Van Den~Berg, Ivan
  Titov, and Max Welling. 2018.
\newblock Modeling relational data with graph convolutional networks.
\newblock In \emph{The Semantic Web: 15th International Conference, ESWC 2018,
  Heraklion, Crete, Greece, June 3--7, 2018, Proceedings 15}, pages 593--607.
  Springer.

\bibitem[{Schwartz et~al.(2015)Schwartz, Reichart, and
  Rappoport}]{schwartz_2015}
Roy Schwartz, Roi Reichart, and Ari Rappoport. 2015.
\newblock Symmetric pattern based word embeddings for improved word similarity
  prediction.
\newblock In \emph{CoNLL}, volume 2015, pages 258--267.

\bibitem[{Shwartz et~al.(2016)Shwartz, Goldberg, and Dagan}]{2016_shwartz}
Vered Shwartz, Yoav Goldberg, and Ido Dagan. 2016.
\newblock Improving hypernymy detection with an integrated path-based and
  distributional method.
\newblock \emph{arXiv preprint arXiv:1603.06076}.

\bibitem[{Van~der Plas and Tiedemann(2006)}]{van_2006}
Lonneke Van~der Plas and J{\"o}rg Tiedemann. 2006.
\newblock Finding synonyms using automatic word alignment and measures of
  distributional similarity.
\newblock In \emph{Proceedings of the COLING/ACL 2006 Main Conference Poster
  Sessions}, pages 866--873.

\bibitem[{Xie and Zeng(2021)}]{xie_2021}
Zhipeng Xie and Nan Zeng. 2021.
\newblock A mixture-of-experts model for antonym-synonym discrimination.
\newblock In \emph{Proceedings of the 59th Annual Meeting of the Association
  for Computational Linguistics and the 11th International Joint Conference on
  Natural Language Processing (Volume 2: Short Papers)}, pages 558--564.

\bibitem[{Yang et~al.(2014)Yang, Yih, He, Gao, and Deng}]{2014_DistMult}
Bishan Yang, Wen-tau Yih, Xiaodong He, Jianfeng Gao, and Li~Deng. 2014.
\newblock Embedding entities and relations for learning and inference in
  knowledge bases.
\newblock \emph{arXiv preprint arXiv:1412.6575}.

\end{thebibliography}
\bibliographystyle{acl_natbib}

\clearpage
\appendix
\section{Appendix}
\label{sec:appendix}

\subsection{Justification for Attentive Convolutions}
\label{appendix_GCN}
In this section, we provide intuitive explanations for:
(a) the limitations posed by the distributional pre-trained word embeddings, 
and 
(b) why attentive graph convolutions are a better choice for capturing 
the relation-specific properties of the data,
(c) computational efficency. 

\paragraph{(a) Word Embeddings.}
We observe the nearest neighbours in the pre-trained word embeddings 
yield a blend of multiple different lexico-semantic relations and 
perform poorly on a specific task.

Underlying reason is the fact that the pre-trained word embeddings 
primarily rely on the distributional hypotheses, i.e., words sharing a 
similar context have similar meanings. From linguistic 
perspective, multiple words with varying relations (i.e., the antonyms 
and synonyms, hypernyms etc.,) may be used interchangeably within a fixed 
context. This in turn results these contextually similar words to be 
embedded close to each other. 
For example, nearest neighbours for the word ``large" in the Glove 
embeddings are a combination of synonyms \{``larger", ``huge"\}, antonyms 
\{``small", ``smaller"\} and irrelevant words \{``sized"\}~\cite{ali2019}.

We argue that in order to refine information from the pre-trained 
embeddings for a specific task, the graphs provide a better 
alternative to analyze the words in combination with semantically 
related neighbours rather than instant-level modeling, as explained below.

\paragraph{(b) Attentive Graph Convolutions.}
The intuitive explanation for the attentive graph convolution network is 
to re-commute the representation of the word via attentive aggregation 
over the neighbouring words.
The core idea is to surround each word by a set of semantically related 
neighbours during the graph construction process in order to
smoothen the representation of the word.

\begin{figure}[t]
    \centering
    \includegraphics[width=0.70\columnwidth]{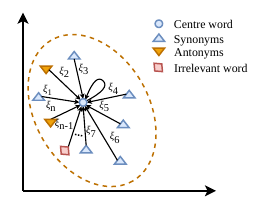}
    \vspace{-2.7ex}
    \caption{Illustration of attentive Graph Convolution Networks}
    \vspace{-2.7ex}
    \label{fig:GCNs}
\end{figure}

It is based on the assumption that within the graphs, i.e., $\text{G}_{t}$ 
and $\text{G}_{h}$, the neighbourhood of each word is dominated by its 
semantically relevant words compared to the antonyms and/or irrelevant words.
And, recomputing the representation of each word by aggregating information 
from the neighbours will result in the final representation to more semantically coherent
compared to the distributional embeddings, as the contribution of antonyms and 
other irrelevant words will be down-weighted.
This is illustrated in Figure~\ref{fig:GCNs}, where the representation of the 
centre word is recomputed using a combination of itself and its nearest 
neighbours (including synonyms, antonyms and irrelevant words). 
We use $\xi_{i}$ as the attentive weight to control its degree of 
association for the i-th neighbor.
The final representation of the word, i.e., the output of the attentive 
graph convolution network is later used for end-task, i.e., antonyms vs synonyms distinction.

\paragraph{(c) Computational Efficiency.}
Another noteworthy aspect is the computational efficiency of the 
attentive graph convolutions.
Theoretically, for each layer the convolutions need to be computed 
between every word pair in the graphs which poses the following limitations: 
(a) it is time consuming and computationally inefficient, 
(b) accumulating information between all possible word pairs may 
incorporate noise in the model training and deteriorate the performance.

To circumvent that we use appropriate thresholds, i.e., $\text{ANT}_{thr}$ and 
$\text{SYN}_{thr}$, to select only highly confident candidates for the 
graph construction. The values for these thresholds are computed empirically.

These thresholds are helpful in cutting down the un-necessary computations 
over the graphs ($\text{G}_{h}$ and $\text{G}_{t}$) by limiting them to the 
neighbourhood $\mathcal{N}_{i}$ of each word $i$. Likewise the attention 
weights between word pairs ($\xi_{i}$) help in cutting down the noise
by appropriately defining the contribution of the neighboring words.
This setting is different from the graph convolution by~\citet{2016_GCN}
that equally consider the contribution of the neighboring nodes in the graph.

\subsection{Difference from R-GCN~\cite{2018_RGCN}}
\label{appndx_RGCN}
\citet{2018_RGCN} is proposed R-GCN, i.e. modeling the Relational data using 
the Graph Convolutional Networks, and used it for entity classification and 
the link prediction task. 
Although, their problem settings for the link prediction task looks similar 
to \OurMODEL, however, we emphasize some key differences as follows:

\begin{enumerate}
    \item R-GCN uses GCN as an encoder to learn the representations 
    followed by DistMult decoder~\cite{2014_DistMult} for link prediction.
    Note, this problem setting is different from ours, as R-GCN 
    primarily deals with asymmetric relations which can be modeled by 
    linear and/or bilinear transformations. On the other hand, \OurMODEL~deals with symmetric 
    relations that cannot be modeled by the existing KG embedding methods, 
    also shown in Figure~\ref{fig:fig1}(b).

    \item Another justification in favour of the above-mentioned argument 
    is the fact that currently the performance of the R-GCN is evaluated 
    on KG embedding data sets, i.e., WN18, FB15k, and these data sets do 
    not include symmetric relation pairs similar to antonym, synonym pairs etc.
    
    \item R-GCN proposes relation-specific feature aggregation for the
    neighbouring nodes via a normalization sum. In contrast, we use 
    attention weights to incorporate the impact of the degree of 
    association of neighboring words/nodes. 
    
    \item \OurMODEL~is the first work that uses multiple encoders to 
    capture the relation-specific properties of the antonym and synonym 
    pairs (i.e., symmetry, transitivity and trans-transitivity), to 
    eventually perform the distinction task in a performance-enhanced way. 
\end{enumerate}

\subsection{Additional Data Sets}
We also test the performance of \OurMODEL~on data sets other than the English.
For this, we used antonym synonym pairs for the Urdu language also used 
by~\citet{ali2019}. 
We acquired this data set from the authors of the Distiller~\cite{ali2019}.
It is a relatively smaller data set encompassing approximately 750 instances, 
priorly splitted into 70\% train, 25\% test and 5\% validation sets. 
For this data set, we used Fasttext embeddings~\cite{2018_Fasttext} for Urdu as 
the pre-trained embeddings. 

The experimental results in Table~\ref{tab:urdu_data} show that 
\OurMODEL~outperforms the baseline models and Distiller by ~\citet{ali2019}
by significant margin. 
Specifically, it improve the F1-score by approximately 3.2\% compared to the existing state-of-the art.
These results also showcase the language-agnostic nature of \OurMODEL. The same 
settings can be applied to the antonyms vs synonyms distinction task for multiple different languages
provided with the availability of distributional embeddings and supervised training data.

\begin{table}[t]
\centering
\resizebox{0.95\columnwidth}{!}{%
\begin{tabular}{l|ccc}
Model & P   & R   & F1  \\
\hline
Baseline-1 (Random Vectors)   & 0.687 & 0.653 & 0.670 \\
Baseline-1 (w/o Graph conv.)  & 0.825 & 0.795 & 0.810 \\
Distiller~\cite{ali2019}      & 0.897 & 0.867 & 0.881 \\
\OurMODEL~                    & \textbf{0.905} & \textbf{0.915} & \textbf{0.910} \\
\hline
\end{tabular}}
\caption{\OurMODEL~performance evaluation using }
\label{tab:urdu_data}
\end{table}

\end{document}